\documentclass[10pt,twocolumn,letterpaper]{article}

\usepackage{iccv}              
\usepackage{multirow}
\usepackage{color}
\usepackage{colortbl}
\definecolor{graycolor}{rgb}{0.95,0.95,0.95}
\usepackage{bm}
\usepackage{siunitx}

\usepackage[labelformat=simple]{subcaption}

\definecolor{iccvblue}{rgb}{0.21,0.49,0.74}
\usepackage[pagebackref,breaklinks,colorlinks,allcolors=iccvblue]{hyperref}

\def\paperID{857} 
\def\confName{ICCV}
\def\confYear{2025}

\title{Learning Dense Feature Matching via Lifting Single 2D Image to 3D Space}

\author{Yingping Liang\textsuperscript{1} \quad 
Yutao Hu\textsuperscript{2} \quad
Wenqi Shao\textsuperscript{3} \quad
Ying Fu\textsuperscript{1$\dagger$} \\
	\textsuperscript{1}Beijing Institute of Technology \qquad \textsuperscript{2}School of Computer Science and Engineering, Southeast University \\ \qquad \textsuperscript{3}Shanghai Al Laboratory\\
	{\tt\small\{liangyingping,fuying\}@bit.edu.cn \quad huyutao@seu.edu.cn \quad shaowenqi@pjlab.org.cn}}

\begin{document}


\twocolumn[{%
\renewcommand\twocolumn[1][]{#1}%
\maketitle

\begin{center}
    \centering
    \includegraphics[width=1.0\textwidth]{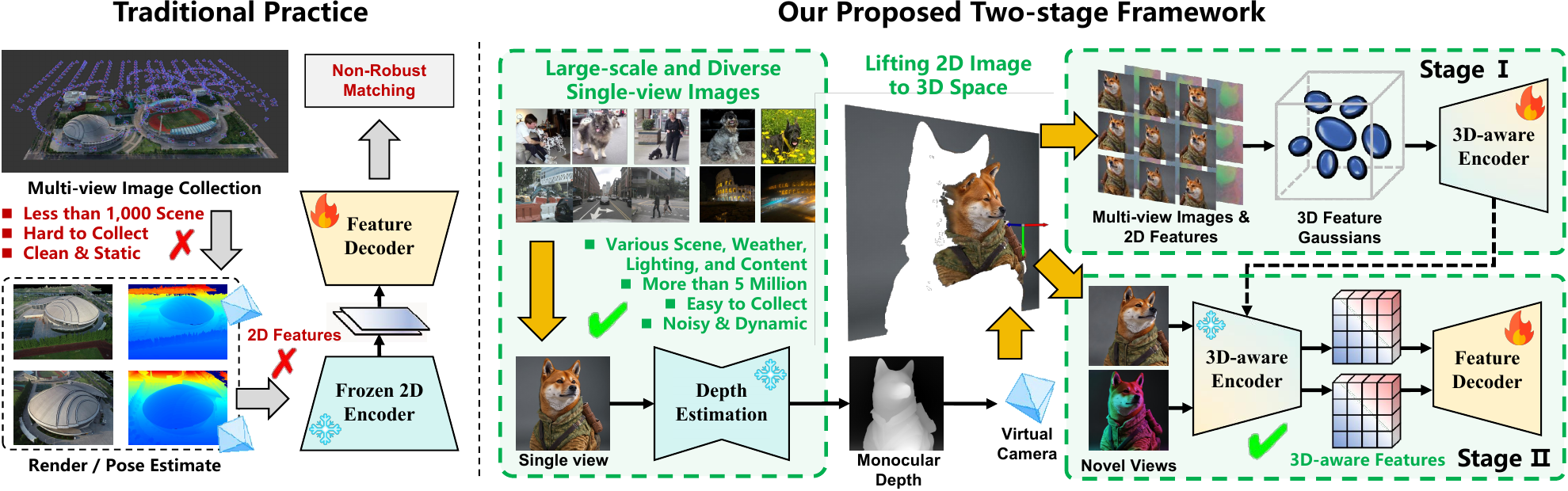}
    \captionof{figure}{Traditional methods rely on multi-view image collections, which are hard to collect and offer limited diversity. Also, encoders are mainly pre-trained on single 2D images and are difficult to capture 3D correspondences in feature matching. One the contrary, our method utilizes large-scale single-view 2D images via lifting them to 3D space and multi-view rendering, providing both 3D-aware encoder trained from 3DGS and robust feature decoder for dense feature matching.}
    \label{fig:teaser}
\end{center}%
}]

{
\renewcommand{\thefootnote}{}
\footnotetext[5]{$\dagger$ Corresponding author.} 
}

\begin{abstract}
   Feature matching plays a fundamental role in many computer vision tasks, yet existing methods heavily rely on scarce and clean multi-view image collections, which constrains their generalization to diverse and challenging scenarios. Moreover, conventional feature encoders are typically trained on single-view 2D images, limiting their capacity to capture 3D-aware correspondences. In this paper, we propose a novel two-stage framework that lifts 2D images to 3D space, named as \textbf{Lift to Match (L2M)}, taking full advantage of large-scale and diverse single-view images. To be specific, in the first stage, we learn a 3D-aware feature encoder using a combination of multi-view image synthesis and 3D feature Gaussian representation, which injects 3D geometry knowledge into the encoder. In the second stage, a novel-view rendering strategy, combined with large-scale synthetic data generation from single-view images, is employed to learn a feature decoder for robust feature matching, thus achieving generalization across diverse domains. Extensive experiments demonstrate that our method achieves superior generalization across zero-shot evaluation benchmarks, highlighting the effectiveness of the proposed framework for robust feature matching. Code is available at \url{https://github.com/Sharpiless/L2M}.
\end{abstract}

\section{Introduction}

Feature matching is a critical task in computer vision, enabling a wide array of applications, including 3D reconstruction \cite{geiger2011stereoscan, luo2024large}, visual localization \cite{sarlin2019coarse, teed2021droid}, and robotics \cite{yang2022fd, tomasi2021learned}. Traditional feature matching methodes, such as SIFT \cite{lowe2004distinctive}, SURF \cite{bay2006surf}, and ORB \cite{rublee2011orb}, primarily rely on hand-crafted descriptors. In recent years, deep learning techniques have significantly advanced feature matching \cite{ma2021image}. Models such as SuperPoint \cite{detone2018superpoint} and DKM \cite{edstedt2023dkm}, have outperformed traditional methods, showing superior to real-world conditions and achieving state-of-the-art results.

However, as shown in Figure \ref{fig:teaser}, current learning-based methods continue to depend heavily on large, annotated 2D image collections \cite{li2018megadepth, yao2020blendedmvs}, typically collected from multi-view cameras and traditional Structure-from-Motion (SfM) algorithms \cite{Schoenberger2016sfm}. These datasets are constrained by the limitations of multi-view 2D image-based datasets, which requires time-consuming multi-view image capture and strict requirements for a static, clean environment. As a result, models trained on such datasets tend to be domain-specific, lacking the generalization ability required to handle diverse scenes and challenging conditions.

A further limitation arises from the design of feature extraction encoders \cite{oquab2023dinov2, he2016deep, dosovitskiy2020vit}, which are typically pre-trained on 2D image datasets, like ImageNet \cite{krizhevsky2012imagenet}, and are optimized to capture 2D features of a single image. However, these 2D features can not incorporate the multi-view perception from different viewpoints \cite{yue2024improving}. Without such 3D geometry knowledge, the encoder struggles to handle occlusions, viewpoint changes, and geometric distortions, leading to unstable matching in complex scenes. Therefore, current feature matching models, which are equipped with such 2D encoders and trained on limited data, struggle to fully establish more reliable matching.

In this paper, we propose a novel two-stage framework, \textbf{Lift to Match (L2M)}, which addresses these limitations by lifting large-scale and diverse 2D images to 3D space. Specifically, \textbf{in the first stage}, to inject 3D geometric knowledge directly into the feature encoder, we propose a novel 3D-aware encoder learning strategy that leverages 3D feature Gaussians to train the feature encoder. Specifically, the encoder is trained on the synthesized multi-view data, guided by explicit 3D features with multi-view perception derived from the 3D feature Gaussians. This enables the encoder to learn multi-view consistent features that are aware of 3D geometry knowledge, rather than just localized 2D textures. The resulting 3D-aware feature encoder is better equipped to handle viewpoint variations, occlusions, and geometric ambiguities. 

Furthermore, \textbf{in the second stage}, we introduce a robust decoder learning strategy, which leverages diverse training data using large-scale single-view images and  novel-view rendering. This learning process enables the feature decoder to produce robust matching results equipped with the frozen 3D-aware encoder. Specifically, by estimating depth from single-view 2D images and reconstructing 3D meshes, we are able to perform novel-view rendering to synthesize large-scale, diverse training data under different lighting conditions. This data generation pipeline enables us to significantly expand the diversity and richness of training samples, covering a wide spectrum of scenes, viewpoints, and lighting conditions. By doing so, L2M breaks free from the domain restrictions of traditional multi-view datasets and enhances the generalization of the trained models.

Experiments demonstrating the state-of-the-art performance of our method across zero-shot evaluation benchmarks. In summary, our main contributions are as follows:

\begin{itemize}
\item We introduce a two-stage framework that lifts 2D images to 3D space for multi-view synthesis and novel-view rendering, which takes advantage of large-scale and diverse single-view images for learning robust feature matching.
\item We propose a 3D-aware encoder learning strategy to adapt 3D geometry knowledge using multi-view synthesis and 3D feature Gaussians, enabling the extracted features to capture multi-view perception.
\item We propose a robust feature decoder learning strategy, which utilizes diverse and large-scale training data via novel-view rendering from single-view 2D images, enhancing the generalization to various scenes.
\end{itemize}

\begin{figure*}[t]
\begin{center}
    \includegraphics[width=0.95\linewidth]{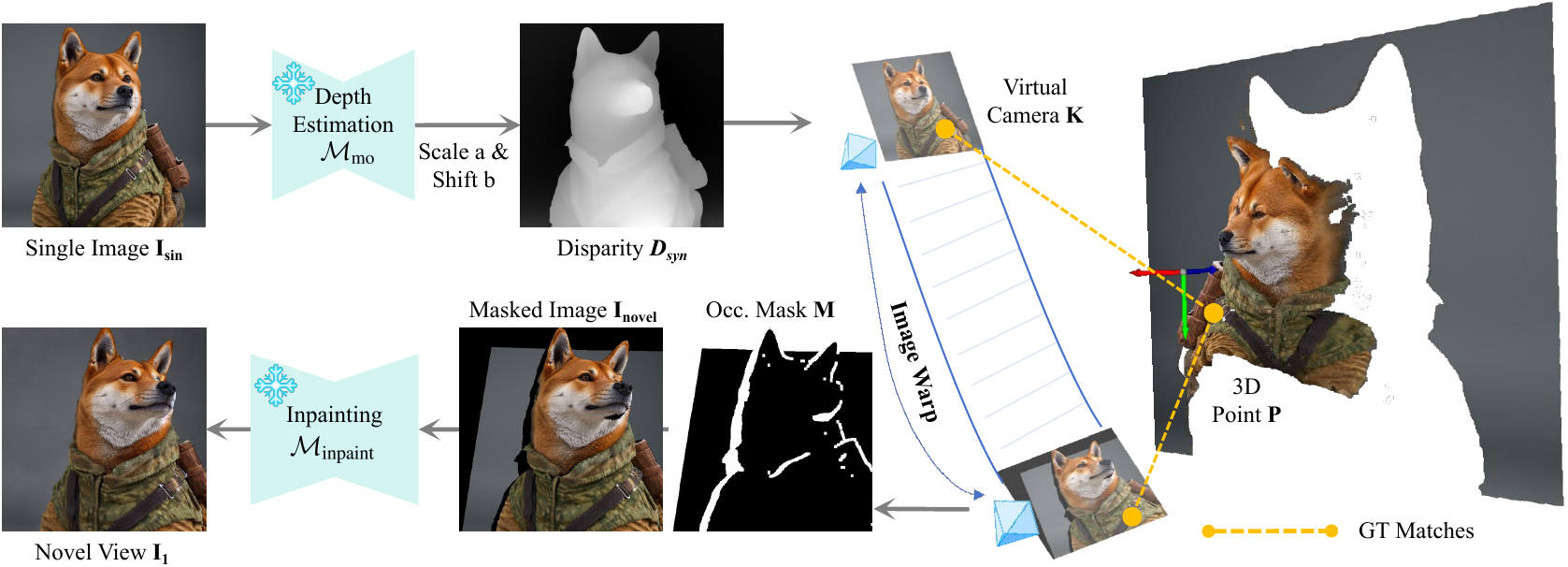} 
\end{center}
\caption{Illustration of our proposed \textbf{novel-view synthesis} strategy via lifting single-view 2D images to 3D space with monocular depth estimation and inpainting, which unlocks the potential for training dense feature matching networks using large-scale, diverse data.}
\label{fig:framework3}
\end{figure*}

\section{Related Work}
\label{sec:related}

\noindent\textbf{Feature Matching Methods.} Feature matching has been a core task in computer vision, with applications spanning from 3D reconstruction to augmented reality and autonomous driving. Early methods primarily relied on handcrafted descriptors, such as SIFT and RootSIFT \cite{arandjelovic2012three}. However, these methods often struggle with lower robustness in real-world scenarios. Recent advances in feature matching have shifted towards learning-based methods. Sparse methods, like SuperGlue \cite{sarlin2020superglue}, leverage deep learning to refine feature matching by modeling spatial relationships. However, they still face challenges in handling variations in lighting and camera. Semi-dense methods such as LoFTR \cite{sun2021loftr} use deep networks to capture long-range dependencies. However, even with these improvements, such methods still struggle with matching under extreme conditions, such as large viewpoint changes or poor texture regions. Dense methods \cite{shen2024gim, edstedt2024roma, edstedt2023dkm} extend feature matching by densely predicting correspondences across the entire image. These methods have shown state-of-the-art results. However, they still face limitations in generalizing across highly complex scenes, especially when trained on limited datasets.

\noindent\textbf{Datasets for Feature Matching.} Current feature matching methods primarily rely on supervised learning, which requires annotated datasets for training. Most publicly available datasets, such as BlendedMVS \cite{yao2020blendedmvs} and Megadepth \cite{li2018megadepth}, focus on small-scale scenarios and fail to capture the full diversity of real-world environments. To overcome these limitations, synthetic data generation has become a popular solution. Techniques such as using game engines \cite{mayer2016large}, forwarding videos \cite{shen2024gim}, and applying 2D affine transformations to single images \cite{bellavia2024image} have been proposed to generate datasets for training. However, such datasets fail to capture the full range of real-world variability, leading to a significant domain gap when applied to real-world data. In contrast to these methods, our method leverages large-scale data generation from real-world single-view images to create diverse training datasets.

\noindent\textbf{Representation Learning.} Vision models often serve as feature extraction encoders for various down-stream tasks. These models, like ResNet \cite{he2016deep} and DINOv2 \cite{oquab2023dinov2}, is often trained on large datasets like ImageNet \cite{krizhevsky2012imagenet} and learn to extract semantic representation from single-view 2D images. However, such models trained on only single-view images focuses on 2D information and may not fully capture the complex 3D geometry knowledge of multi-view images needed for accurate feature matching across different views. Fit3D~\cite{yue2024improving} proposes the use of multi-view 3D Gaussians collections to fine-tune 2D feature representations, but is still suffering from hard-to-collect multi-view images. To address this gap, we introduce a learning process to incorporate 3D geometry knowledge into the encoders, which requires only single-view 2D images.

\begin{figure*}[t]
\begin{center}
    \includegraphics[width=1.0\linewidth]{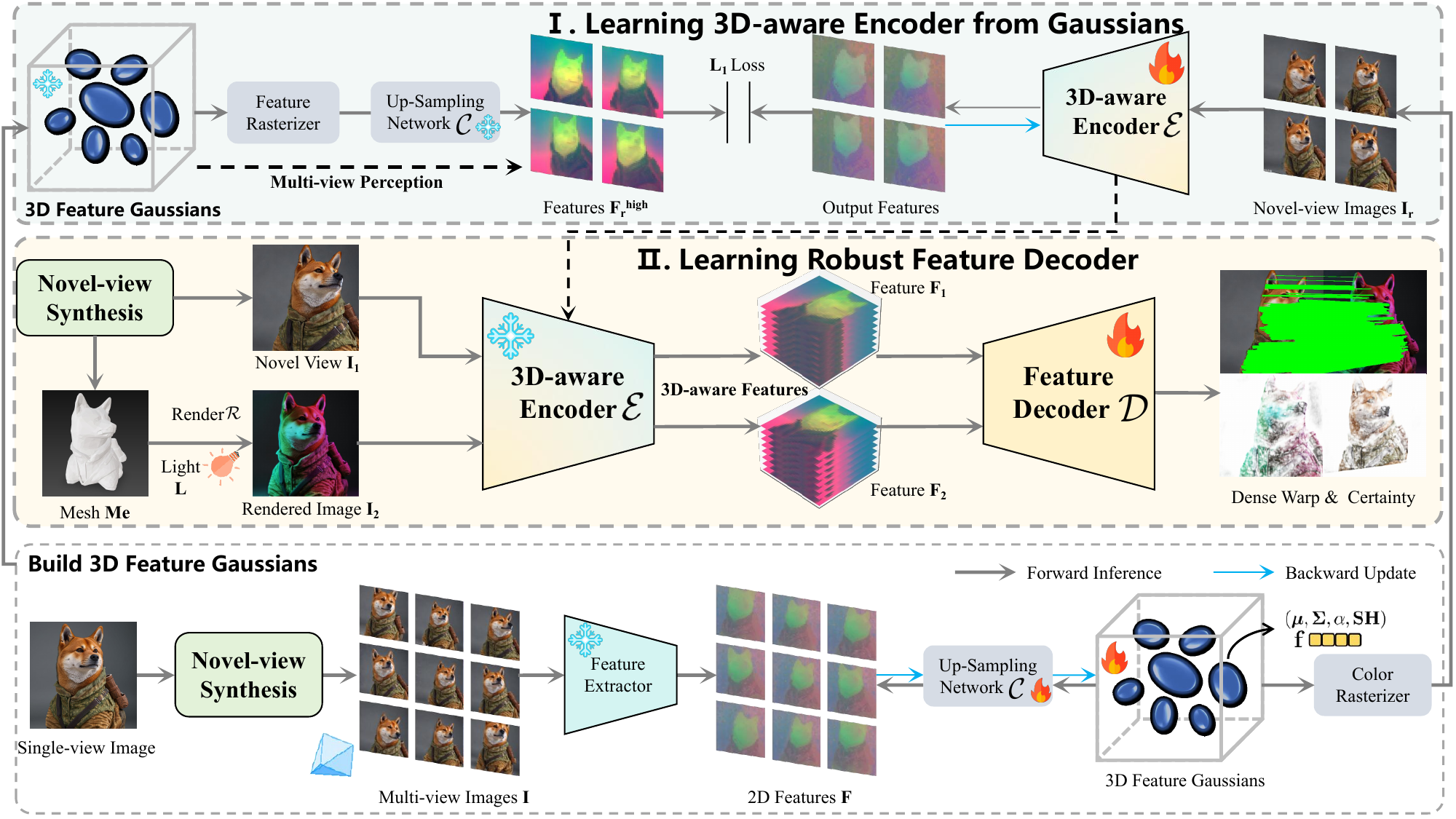} 
\end{center}
\caption{Illustration of our two-stage framework. In the first stage, the 3D-aware feature encoder learning process utilizes multi-view synthesis and 3D feature Gaussians to transfer 3D geometry knowledge into the encoder. In the second stage, the robust feature decoder learning process utilizes large-scale, easy-to-collect single views with re-rendering strategy, providing much more diverse data for training.}
\label{fig:framework2}
\end{figure*}

\section{Method}

In this section, we first detail the formulation and motivation, as well as the novel-view synthesis strategy via lifting single-view 2D images to 3D space, as shown in Figure \ref{fig:framework3}. Then, as shown in Figure~\ref{fig:framework2}, we introduce the 3D-aware encoder learning process with 3D feature Gaussians. After that, we describe the robust decoder learning process. Finally, we provide the implementation details.

\subsection{Formulation and Motivation}

In dense feature matching, given two input images \( \mathbf{I}_{1} \) and \( \mathbf{I}_{2} \), we first extract their feature representations using a shared encoder:
\begin{eqnarray}
    \mathbf{F}_{1} = \mathcal{E}(\mathbf{I}_{1}), \mathbf{F}_{2} = \mathcal{E}(\mathbf{I}_{2}),
\end{eqnarray}
where $\mathcal{E}$ is the feature encoder with shared weights. These features are then passed to a decoder, which predicts the pixel-wise transformation (warp) \( \mathbf{W} \) and certainty \( \mathbf{\sigma} \):
\begin{eqnarray}
    \{\mathbf{W}, \mathbf{\sigma}\} = \mathcal{D}(\mathbf{F}_{1}, \mathbf{F}_{2}).
\end{eqnarray}

However, there are still two main challenges. First, state-of-the-art feature matching models rely on 2D vision encoders. These encoders are typically trained on single 2D images and are not capable of capturing 3D geometry knowledge, which limits their performance in complex or dynamic environments. We overcome this limitation by training a 2D vision model into a 3D-aware encoder, which injects multi-view perception into the feature extraction process with the help of 3D feature Gaussians.

Second, collecting large-scale, diverse training data is difficult and also expensive, as multi-view image datasets that cover various domains and conditions are both costly and labor-intensive, which restrict their generalization across different real-world scenarios. Our framework addresses this by generating large-scale, diverse datasets using single-view depth estimation and novel-view rendering.


\subsection{Lifting 2D Image to 3D for Novel-view Synthesis}

Specifically, to lift 2D image to 3D space, we first use a pre-trained monocular depth estimation model, such as Depth Anything V2 \cite{yang2025depth}, which predicts depth maps from single RGB images. For each natural image \( \mathbf{I}_{\text{sin}} \), we use a monocular depth estimation model to predict the dense depth map \( \mathbf{D}_{\text{syn}} \) and sample a random scale $a$ and shift $b$:
\begin{eqnarray}
    \mathbf{D}_{\text{syn}} = a \times \mathcal{M}_{\text{mo}}(\mathbf{I}_{\text{sin}}) + b,
\end{eqnarray}
where \( \mathcal{M}_{\text{mo}} \) represents the monocular depth estimation model. These synthesized depth maps, though not accurate in metric scale, capture the relative depth relationships and structural details in the scene, providing valuable supervision signals during pre-training.

Then we use the predicted depth to lift the single-view image into 3D space. We first sample a random camera intrinsic matrix \( \mathbf{K} \). Next, for each pixel \( (u, v) \) in the depth map, we compute the corresponding 3D coordinates in the camera coordinate system using the sampled camera intrinsic matrix \( K \) and the depth value at that pixel. This transformation generates a point cloud \( \mathbf{P} = \{(X, Y, Z)\} \) representing the 3D spatial locations of each pixel.

We then warp the image to render novel view images from new perspectives, applying masks to account for occlusions. Specifically, the mask \( \mathbf{M} \) is used to indicate which parts of the image are visible and which are occluded. To handle occlusions, we use an inpainting model \( \mathcal{M}_{\text{inpaint}} \) to fill in the missing regions in the rendered images. The inpainting process can be represented as:
\begin{eqnarray}
\mathbf{I}_{1} = \mathcal{M}_{\text{inpaint}}(\mathbf{I}_{\text{novel}}, \mathbf{M}),
\end{eqnarray}
where \( \mathcal{M}_{\text{inpaint}} \) is the inpainting model that reconstructs the occluded parts of the image based on the visible regions. This process generates paired images with corresponding depth maps and camera parameters, providing valuable training data for dense feature matching models.

\subsection{Learning 3D-aware Encoder from Gaussians}

Traditional feature encoders are typically designed to extract 2D features, which are insufficient for capturing the full 3D structure and multi-view perception necessary for accurate feature matching. To address these limitations, we combine the multi-view generation and 3D feature Gaussians, which incorporates 3D geometry knowledge into the feature encoders. This process allows the feature encoder to better understand multi-view perception.

\noindent\textbf{Building 3D Feature Gaussians.} Utilizing the multi-view generation method, we are able to generate a set of multi-view images \( \{\mathbf{I}_i\}_{1 \leq i \leq N} \) and corresponding feature maps \( \{\mathbf{F}_i\}_{1 \leq i \leq N} \) from a 2D feature extraction encoder (e.g., DINOv2 \cite{oquab2023dinov2}). These feature maps are then used to build the 3D feature Gaussians. 

The goal for building 3D feature Gaussians is to optimize the Gaussian parameters such that both the images \( \mathbf{I} \) and feature maps \( \mathbf{F} \) are well-represented in the 3D space, aligning the 2D features with the 3D structure. Following \cite{yue2024improving}, a set of 3D Gaussians is defined as:
\begin{eqnarray}
\mathcal{G} = \{(\bm{\mu}, \mathbf{s}, \mathbf{R}, \alpha, \mathbf{SH}, \mathbf{f})_j\}_{1 \leq j \leq M},
\end{eqnarray}
where \( \bm{\mu} \) is the 3D mean, \( \mathbf{s} \) is the scale, \( \mathbf{R} \) is the orientation, and \( \alpha \) is opacity. Additionally, \( \mathbf{SH} \) represents view-dependent color, and \( \mathbf{f} \) stores the distilled 2D features in 3D space. In order to reduce the computational cost, a trainable CNN $\mathcal{C}$ is used to reduce the dimension of the features. 

\noindent\textbf{Learning 3D-aware Encoder.} After optimizing the 3D feature Gaussian parameters for the scene, we can render from the Gaussians a set of novel-view images $\mathbf{I}_{\mathrm{r}}$ and the low-dimension feature maps $\mathbf{F}_{\mathrm{r}}^{\mathrm{low}}$. To be specific, the images and features can be rendered using a differentiable feature rasterizer, based on an $\alpha$-blending method:
\begin{eqnarray}
\mathbf{F}_{\mathrm{r}}^{\mathrm{low}} = \sum_{i\in \mathcal{N}} \mathbf{f}_{i} \alpha_{i} \prod_{j=1}^{i-1} (1-\alpha_{i}),
\end{eqnarray}
where \( \mathcal{N} \) is the set of overlapping Gaussians, and \( \alpha_{i} \) is the opacity evaluated from the Gaussian's covariance matrix. This process produces low-dimensional feature images, which are then up-projected to higher dimensions using the CNN-based Up-sampling network: $\mathbf{F}_{\mathrm{r}}^{\mathrm{high}} = \mathcal{C} (\mathbf{F}_{\mathrm{r}}^{\mathrm{low}})$. These feature maps are then used to train the encoder by a pixel-wise $L_1$ loss. This process enables the encoder to better capture 3D geometry knowledge.

\begin{figure}[t]
\begin{center}
    \includegraphics[width=1.0\linewidth]{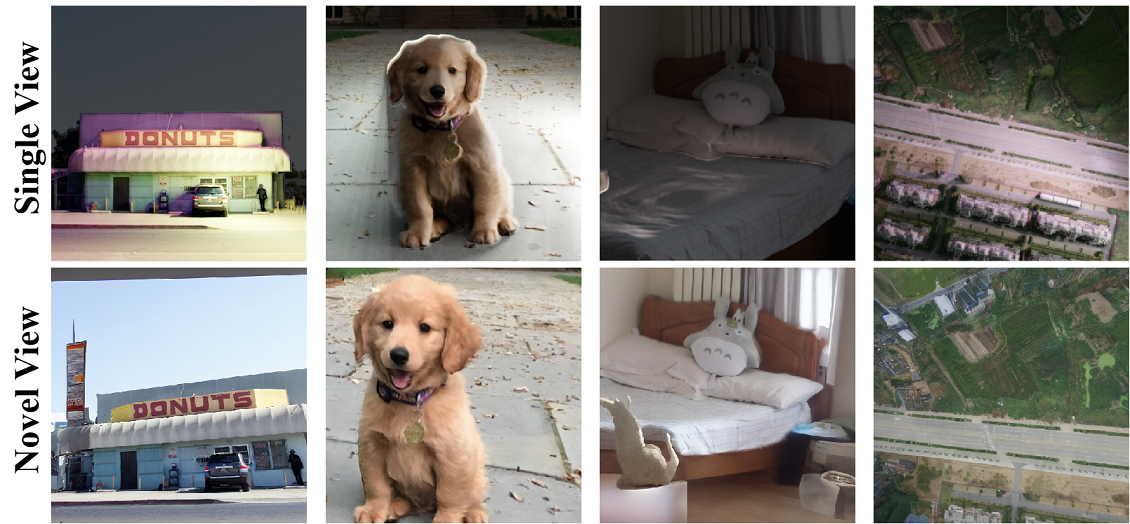} 
\end{center}
\caption{Example of generated image pairs. The first row shows the original single-view images after re-lighting and re-rendering. The second row shows the generated novel-view images.}
\label{fig:visualization1}
\end{figure}

\subsection{Learning Robust Feature Decoder}
\label{sec:data-generation}

While the first stage focuses on enhancing the feature encoder with 3D awareness, the second stage aims to learn a robust feature matching decoder that can generalize across diverse image pairs, including those with significant viewpoint, lighting, and appearance variations. A critical challenge in this stage lies in the scarcity of large-scale multi-view training data, which traditionally requires labor-intensive collection of calibrated image pairs with known camera poses and depths.

To overcome this challenge, we design a scalable data generation pipeline that leverages monocular depth estimation to synthesize diverse training pairs from single-view images. This pipeline allows us to construct training data without requiring explicit multi-view supervision, significantly broadening the domain coverage of the training set.

To be specific, we take two ways to get the images $\mathbf{I}_{1}$ and $\mathbf{I}_{2}$, separately. First, the data generation pipeline begins with a single-view image $\mathbf{I}_{sin}$. The image $\mathbf{I}_{1}$ is synthesized by a novel view synthesis strategy, combining monocular depth estimation, image warping and inpainting techniques. Then, the monocular depth and re-light technique are used to obtain the image $\mathbf{I}_{2}$ under a novel lighting condition.

\begin{table}[t] 
\small
\centering
\caption{Real-world datasets with diverse single-view images we used for training data generation.}
\begin{tabular}{@{}lcccc@{}}
\toprule
\textbf{Dataset} & \textbf{Indoor} & \textbf{Outdoor} & \# \textbf{Images} & \textbf{Scene} \\ \midrule
COCO \cite{lin2014microsoft} & \checkmark & \checkmark & 118,287 & Common \\
DAVIS \cite{perazzi2016benchmark} & \checkmark & \checkmark & 10,581 & Common \\
ADE20K \cite{zhou2019semantic} & \checkmark & \checkmark & 19,983 & Common \\
GLDv2 \cite{weyand2020google} & & \checkmark & 117,576 & Landmarks \\
Nuscenes \cite{caesar2020nuscenes} & & \checkmark & 93,475 & Urban \\
Cityscapes \cite{cordts2016cityscapes} & & \checkmark & 19,998 & Urban \\ 
KITTI \cite{Geiger2012CVPR} & & \checkmark & 93,657 & Urban \\
LOL \cite{Chen2018Retinex} & \checkmark & & 500 & Low-light \\
LOLI \cite{li2021low} & \checkmark & \checkmark & 200 & Low-light \\
NYU V2 \cite{silberman2012indoor} & \checkmark & & 45,205 & Indoor \\
LSUI \cite{peng2023u} & & \checkmark & 5,004 & Underwater  \\
UAV \cite{xu2024uav} & & \checkmark & 1,359 & Aerial \\
\bottomrule
\end{tabular}
\label{tab:datasets}
\end{table}

\begin{table*}[h]
\small
    \centering
    \caption{Comparison of different methods on Zero-shot Evaluation Benchmark (ZEB) \cite{shen2024gim}, which consists of 12 public datasets that cover a variety of scenes and conditions. The AUC of the pose error under 5° (\%) is reported. In this table, we mainly compare our method with all dense methods, which present the state-of-the-art and show significant advantages over sparse and semi-dense methods. We also show the results of representative sparse and semi-dense methods to provided broader context.}
    \setlength{\tabcolsep}{1.2mm}
    \resizebox{\textwidth}{!}{
    \begin{tabular}{l l c c c c c c c c c c c c c c}
        \toprule
        \multirow{2}{*}{\textbf{Category}} & \multirow{2}{*}{\textbf{Method}} & \multirow{2}{*}{\textbf{Mean}} & \multicolumn{8}{c}{\textbf{Real-world Datasets}} & \multicolumn{4}{c}{\textbf{Synthetic Datasets}} \\ \cmidrule(lr){4-11} \cmidrule(lr){12-15}
        & & & GL3 & BLE & ETI & ETO & KIT & WEA & SEA & NIG & MUL & SCE & ICL & GTA \\
        \midrule
        \textbf{Handcrafted}
        & RootSIFT \cite{arandjelovic2012three} & 31.8 & 43.5 & 33.6 & 49.9 & 48.7 & 35.2 & 21.4 & 44.1 & 14.7 & 33.4 & 7.6 & 14.8 & 35.1 \\
        \midrule
        \multirow{3}{*}{\textbf{Sparse}}
        & SuperGlue (indoor) \cite{sarlin2020superglue} & 21.6 & 19.2 & 16.0 & 38.2 & 37.7 & 22.0 & 20.8 & 40.8 & 13.7 & 21.4 & 0.8 & 9.6 & 18.8 \\
        & SuperGlue (outdoor) \cite{sarlin2020superglue} & 31.2 & 29.7 & 24.2 & 52.3 & 59.3 & 28.0 & 28.4 & 48.0 & 20.9 & 33.4 & 4.5 & 16.6 & 29.3 \\
        & LightGlue \cite{lindenberger2023lightglue} & 31.7 & 28.9 & 23.9 & 51.6 & 56.3 & 32.1 & 29.5 & 48.9 & 22.2 & 37.4 & 3.0 & 16.2 & 30.4 \\
        \midrule
        \multirow{4}{*}{\textbf{Semi-Dense}}
        & LoFTR (indoor) \cite{sun2021loftr} & 10.7 & 5.6 & 5.1 & 11.8 & 7.5 & 17.2 & 6.4 & 9.7 & 3.5 & 22.4 & 1.3 & 14.9 & 23.4 \\
        & LoFTR (outdoor) \cite{sun2021loftr} & 33.1 & 29.3 & 22.5 & 51.1 & 60.1 & 36.1 & 29.7 & 48.6 & 19.4 & 37.0 & 13.1 & 20.5 & 30.3 \\
        & ELoFTR (outdoor) \cite{wang2024efficient} & 32.8 & 27.7 & 22.8 & 50.7 & 62.7 & 35.9 & 28.1 & 46.1 & 16.7 & 38.1 & 12.2 & 22.7 & 30.0 \\
        \midrule
        \multirow{6}{*}{\textbf{Dense}}
        & DKM (indoor) \cite{edstedt2023dkm} & 46.2 & 44.4 & 37.0 & 65.7 & 73.3 & 40.2 & 32.8 & 51.0 & 23.1 & 54.7 & 33.0 & \underline{43.6} & 55.7 \\
        & DKM (outdoor) \cite{edstedt2023dkm} & 45.8 & 45.7 & 37.0 & 66.8 & 75.8 & 41.7 & 33.5 & 51.4 & 22.9 & 56.3 & 27.3 & 37.8 & 52.9 \\
        & GIM \cite{shen2024gim} & \underline{51.2} & \textbf{63.3} & \textbf{53.0} & \underline{73.9} & 76.7 & \underline{43.4} & \underline{34.6} & \underline{52.5} & \underline{24.5} & 56.6 & 32.2 & 42.5 & \textbf{61.6} \\
        & RoMa (indoor) \cite{edstedt2024roma} & 46.7 & 46.0 & 39.3 & 68.8 & 77.2 & 36.5 & 31.1 & 50.4 & 20.8 & 57.8 & \underline{33.8} & 41.7 & 57.6 \\
        & RoMa (outdoor) \cite{edstedt2024roma} & 48.8 & 48.3 & 40.6 & 73.6 & \underline{79.8} & 39.9 & 34.4 & 51.4 & 24.2 & \underline{59.9} & 33.7 & 41.3 & 59.2 \\
        \rowcolor{graycolor} \cellcolor{white} & \textbf{L2M (Ours)} & \textbf{51.8} & \underline{51.5} & \underline{46.0} & \textbf{77.2} & \textbf{83.7} & \textbf{44.9} &   \textbf{36.0} & \textbf{52.9} & \textbf{25.3} & \textbf{61.7} & \textbf{38.5} &  \textbf{43.8} & \underline{60.6} \\
        \bottomrule
    \end{tabular}
    }
    \label{tab:results}
\end{table*}

Furthermore, we fully leverage the capabilities of the physics engine to re-render the original mesh from different viewpoints, simulating diverse conditions such as varying lighting. Using the 3D point cloud from novel-view synthesis process, we can reconstruct a 3D mesh $\mathbf{Me}$ through surface reconstruction techniques, such as Poisson Surface Reconstruction \cite{kazhdan2006poisson}, to create a continuous 3D surface model for re-rendering. Then, to simulate different lighting conditions, we introduce a light source vector \( \mathbf{L} \) and modify the rendering equation to account for lighting variations:
\begin{eqnarray}
\mathbf{I}_{2} = \mathcal{R}(\mathbf{Me}, \mathbf{L}),
\end{eqnarray}
where \( \mathcal{R} \) is the rendering function that takes the mesh \( \mathbf{Me} \) and light source \( \mathbf{L} \) into account. 

Then, the paired images $\mathbf{I}_{1}$ and $\mathbf{I}_{2}$ with dense matching labels can be used to train the feature decoder equipped with our 3D-aware feature encoder. This allows us to generate a diverse set of images, which improves the model's robustness and enables it to generalize to unseen scenarios.

\subsection{Implementation Details}

\noindent\textbf{Data Sources.} To generate diverse training data, we leverage a range of rich, real-world datasets containing single-view images, as shown in Table \ref{tab:datasets}. These datasets cover both indoor and outdoor environments, providing a variety of scenes and conditions to ensure the generalizability of the learned models across different domains. The datasets used for this purpose include COCO \cite{lin2014microsoft}, Google Landmarks \cite{weyand2020google}, Nuscenes \cite{caesar2020nuscenes}, Cityscapes \cite{cordts2016cityscapes}, and others, which offer a combination of urban, natural, and indoor scenes, along with variations in lighting, objects, and cameras. 

\noindent\textbf{Training Parameters.} We use a canonical learning rate (for batchsize = 8 per GPU) of $10^{-4}$ for the decoder, and $5\times10^{-6}$ for the encoder. The models are trained on a resolution of $584\times584$. The training process takes about 3.5 days on 4 A100 80GB GPUs. For inpainting model, we use Stable-Diffusion v1.5 \cite{rombach2022high}. For encoder fine-tuning, we randomly sample 10,000 images and synthesize 9 novel views per image. For decoder training, we use all the images (around 525,000 in total) and generate one image pair from each image. The focal length in the camera intrinsic matrix $K \in [0.58, 0.88]$. The lighting conditions are varied by randomly changing the number (1–3), intensity (1000–3000), color, and position. For 3DGS construction, we follow the setup in FiT3D \cite{yue2024improving}.

\begin{table}
\small
    \centering
    \caption{Performance comparison on MegaDepth-1500 when trained or fine-tuned on the MegaDepth training set.}
    \begin{tabular}{l lccc }
    \toprule
    \multirow{2}{*}{\textbf{Category}} & \multirow{2.3}{*}{\textbf{Method}} & \multicolumn{3}{c}{\textbf{Pose estimation AUC}}\\
    \cmidrule(lr){3-5}
    & & \textbf{@$5^\circ$} & \textbf{@$10^\circ$} & \textbf{@$20^\circ$} \\
     \midrule
       \multirow{2}{*}{\textbf{Sparse}} & SuperGlue \cite{sarlin2020superglue} & 42.2 & 61.2 & 76.0 \\
        & LightGlue~\cite{lindenberger2023lightglue}  & 51.0 & 68.1 & 80.7\\
         \midrule
         \multirow{5}{*}{\textbf{Semi-Dense}} & LoFTR~\cite{sun2021loftr} & 52.8 & 69.2 & 81.2 \\
         & ELoFTR \cite{wang2024efficient} & 56.4 & 72.2 & 83.5 \\
         & XFeat~\cite{potje2024xfeat} & 50.2 & 65.4 & 77.1 \\
         & ASpanFormer~\cite{chen2022aspanformer} & 55.3  & 71.5 & 83.1\\
         & ASTR~\cite{yu2023adaptive} & 58.4 & 73.1 & 83.8\\
         \midrule
         \multirow{4}{*}{\textbf{Dense}} & DKM~\cite{edstedt2023dkm}  &  60.4 & 74.9 & 85.1 \\
         & GIM \cite{shen2024gim}  &  60.7 & 75.5 & 85.9 \\
         & RoMa \cite{edstedt2024roma} & 
         \underline{62.6} & \underline{76.7} & \underline{86.3} \\
         \rowcolor{graycolor} \cellcolor{white} & \textbf{L2M (Ours)} & \textbf{63.1} & \textbf{77.1} & \textbf{86.6} \\
    \bottomrule
    \end{tabular}
    \label{tab:megadepth}
\end{table}


\section{Experiments}

In this section, we first introduce the datasets and evaluation metrics for experiments. Then, detailed comparisons are conducted with the state-of-the-art methods. Finally, ablations and discussions are performed to confirm the effectiveness of the main components. Additional experiments and analysis are provided in the supplementary materials.

\subsection{Evaluation Datasets and Metrics}

\noindent\textbf{Evaluation Datasets.} To analyze the robustness of our models on in-the-wild data, we use a comprehensive zero-shot evaluation benchmark (ZEB) \cite{shen2024gim}, which includes 8 real-world datasets and 4 simulated datasets with diverse image resolutions, scene conditions and view points. We also evaluate the zero-shot performance of our methods on the in-domain dataset after fine-tuning on MegeDepth dataset \cite{li2018megadepth} and the cross-modal performance for RGB-IR matching on METU-VisTIR \cite{tuzcuouglu2024xoftr} dataset.

\begin{table}[t]
\small
\centering
\caption{Zero-shot performance comparison on RGB-IR Dataset (METU-VisTIR \cite{tuzcuouglu2024xoftr}). The AUC of the pose error (\%) is reported. ``*" indicates cross-modal methods.}
\begin{tabular}{l l c c c }
\toprule
\multirow{2}{*}{\textbf{Category}} & \multirow{2.3}{*}{\textbf{Method}} & \multicolumn{3}{c}{\textbf{Pose estimation AUC}}\\
\cmidrule(lr){3-5}
& & \textbf{@$5^\circ$} & \textbf{@$10^\circ$} & \textbf{@$20^\circ$} \\
\midrule
\multirow{3}{*}{\textbf{Sparse}}
    & SuperGlue \cite{sarlin2020superglue} & 4.30 & 9.26 & 17.21 \\
    & LightGlue \cite{lindenberger2023lightglue} & 2.17		& 5.37		& 11.21 \\
    & ReDFeat* \cite{deng2022redfeat} & 1.71	 & 4.57	 & 10.85 \\
\midrule
\multirow{7}{*}{\textbf{Semi-Dense}}
    & LoFTR \cite{sun2021loftr} & 2.88  &	 6.94	 & 14.95 \\
    & ELoFTR \cite{wang2024efficient} & 2.88 & 	7.88 & 	17.72 \\
    & XFeat~\cite{potje2024xfeat} & 2.35 & 6.08 & 14.45 \\
    & ASpanFormer~\cite{chen2022aspanformer} & 2.47 & 5.86 & 12.39 \\
    & CasMTR~\cite{cao2023casmtr} & 3.12 & 5.50 & 18.89 \\
    & XoFTR* \cite{tuzcuouglu2024xoftr} & 18.47 & 34.64 & 51.50 \\
\midrule
\multirow{4}{*}{\textbf{Dense}}
    & DKM \cite{edstedt2023dkm}  & 6.76 & 13.69 & 22.53 \\
    & GIM \cite{shen2024gim} & 5.08 & 12.30 & 23.69 \\
    & RoMa \cite{edstedt2024roma} & \underline{25.61} & \underline{48.12} & \underline{68.37} \\
   \rowcolor{graycolor} \cellcolor{white} & \textbf{L2M (Ours)} & \textbf{30.13} & \textbf{53.11} & \textbf{71.80} \\
\bottomrule
\end{tabular}
\label{tab:Real_IR_pose} 
\end{table}

\begin{figure*}[t]
    \centering
    \subfloat[Image Pair]{\includegraphics[width=1.3in]{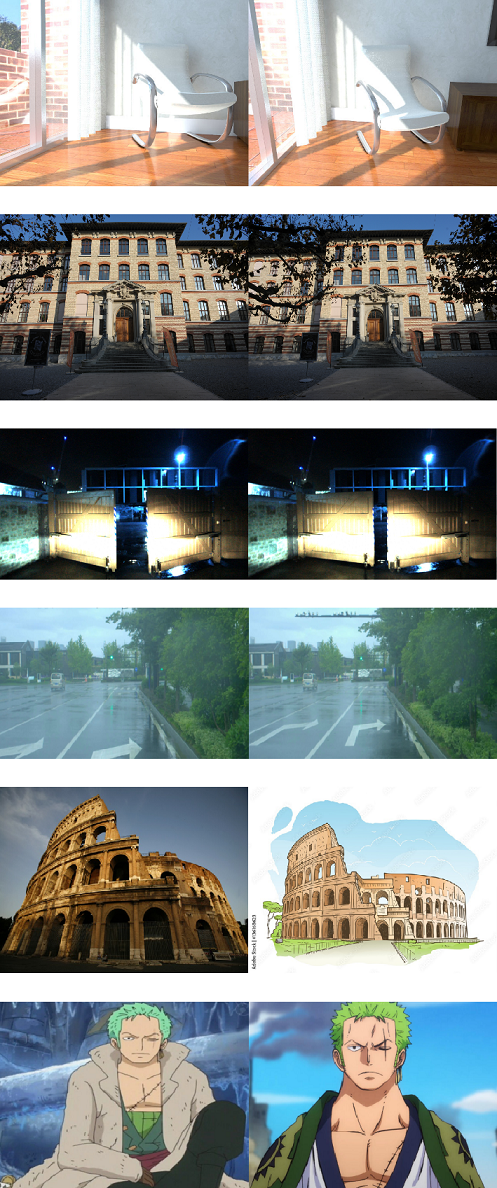}}
    \hfil
    \subfloat[DKM \cite{edstedt2023dkm}]{\includegraphics[width=1.3in]{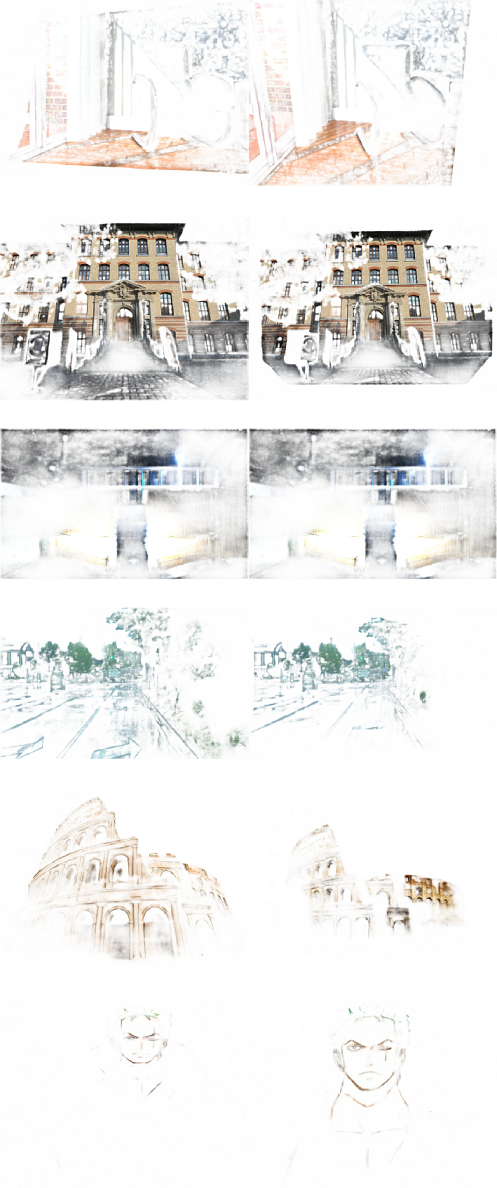}}
    \hfil
    \subfloat[GIM \cite{shen2024gim}]{\includegraphics[width=1.3in]{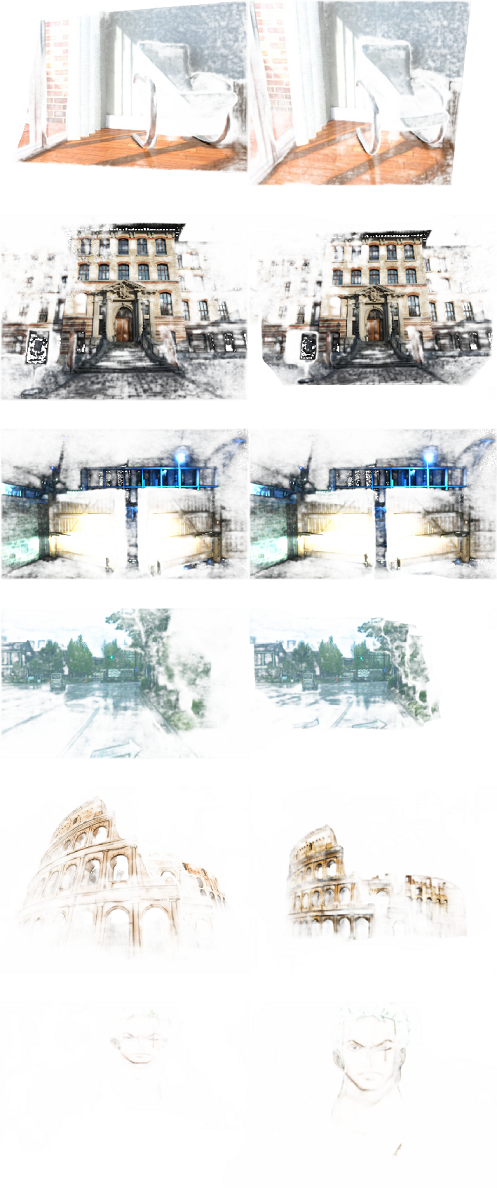}}
    \hfil
    \subfloat[ROMA \cite{edstedt2024roma}]{\includegraphics[width=1.3in]{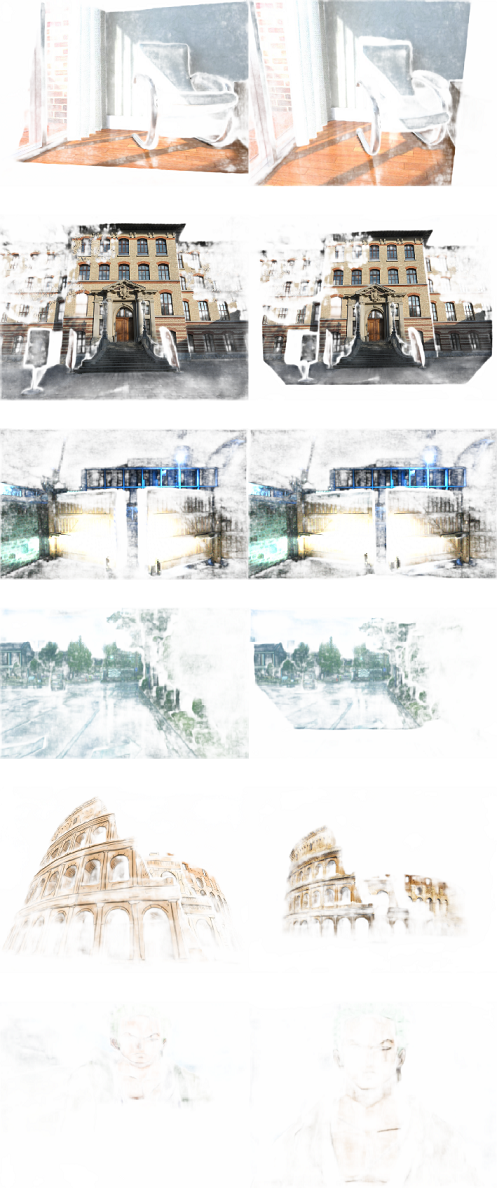}}
    \hfil
    \subfloat[L2M (Ours)]{\includegraphics[width=1.3in]{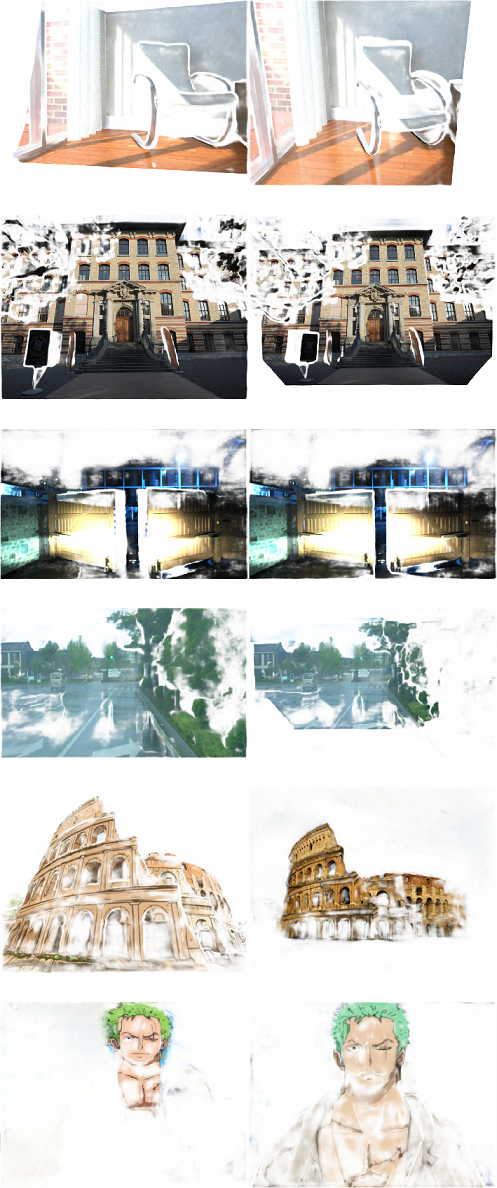}}
    \caption{Qualitative comparison with dense feature matching methods \cite{edstedt2023dkm, shen2024gim, edstedt2024roma}, which represent the state-of-the-art and can output dense pixel-to-pixel results. We show the results of warp $\times$ certainty, under different weather, lighting, and style conditions. The results indicate that our proposed method can establish much more precise correspondences and denser matching results.}
    \label{fig:visualization_warp}
\end{figure*}

\noindent\textbf{Evaluation Metrics.} For evaluation metrics on RGB datasets, following GIM \cite{shen2024gim}, we report the AUC of the relative pose error within $5^{\circ}$, where the pose error is the maximum between the rotation angular error and translation angular error. The relative poses are obtained by estimating the essential matrix using the output correspondences from the matching methods and RANSAC. For cross-modal datasets, the recovered poses by matches are evaluated to measure the accuracy. We report the area under the curve (AUC) of the pose error at thresholds {$5^{\circ}$, $10^{\circ}$, $20^{\circ}$}.

\subsection{Main Results}

In this work, we primarily focus on comparing against dense feature matching methods, as they represent the current state-of-the-art in feature matching research. We also report results for several representative sparse and semi-dense methods to provide broader context.

\noindent\textbf{Zero-shot Performance Evaluation.} As shown in Table \ref{tab:results}, we present a comprehensive comparison on the Zero-shot Evaluation Benchmark (ZEB) \cite{shen2024gim}, which consists of \textbf{12} public datasets covering a variety of scenes and weather conditions. The benchmark includes both real-world datasets and synthetic datasets, with the performance measured by the AUC of pose errors at a threshold of \(5^\circ\). Note that, ``outdoor" indicates models trained on MegaDepth and ``indoor" indicates models trained on both MegaDepth and Scannet. Note that ELoFTR \cite{wang2024efficient} does not provide the indoor checkpoints. Our method consistently outperforms other techniques on most cases. Notably, we achieve the highest AUC values on several challenging datasets such as SEA (\(\textbf{52.9\%}\)) and WEA (\(\textbf{32.0\%}\)). Our performance remains robust even in more challenging settings, outperforming other methods. This results confirming its ability to generalize well across real-world conditions.

\begin{table*}[h]
    \small
    \centering
    \caption{Ablation study on the main components: 1) incorporating the 3D-aware encoder (Stage 1), and 2) utilizing large-scale and diverse synthetic data for training the decoder (Stage 2). We only use MegaDepth dataset for training when not using the data from Stage 2.}
    \begin{tabular}{l c c c c c c c c c c c c c c}
        \toprule
        \multirow{2}{*}{\textbf{Method}} & \multicolumn{8}{c}{\textbf{Real-world Datasets}} & \multicolumn{4}{c}{\textbf{Synthetic Datasets}} \\ \cmidrule(lr){2-9} \cmidrule(lr){10-13}
        & GL3 & BLE & ETI & ETO & KIT & WEA & SEA & NIG & MUL & SCE & ICL & GTA \\
        \midrule
        \rowcolor{graycolor} \textbf{L2M} & \textbf{51.5} & \textbf{46.0} & \textbf{77.2} & \textbf{83.7} & \textbf{44.9} &   \textbf{36.0} & \textbf{52.9} & \textbf{25.3} & \underline{61.7} & \textbf{38.5} &  \textbf{43.8} & \textbf{60.6} \\
         \quad w/o Stage 1 & \underline{50.2} & \underline{41.6} & \underline{75.4} & \underline{83.6} & \underline{42.9} &   \underline{35.2} & \underline{52.5} & \underline{25.2} & \textbf{61.9} & \underline{34.4} & \underline{41.9} & \underline{59.5} \\
         \quad w/o Stage 1 \& Stage 2 & 46.0 & 39.3 & 68.8 & 77.2 & 36.5 & 31.1 & 50.4 & 20.8 & 57.8 & 33.8 & 41.7 & 57.6 \\
        \bottomrule
    \end{tabular}
    \label{tab:ablations}
\end{table*}

\noindent\textbf{In-domain Performance Evaluation.} We also evaluate the im-domain performance of our method on the MegaDepth-1500 test set \cite{sun2021loftr} when fine-tuned on MegaDepth training set. The test set includes 1500 image pairs with variable weather, occlusion, and lighting conditions from two challenging scenes: scene 0015 and scene 0022. Following the protocol from \cite{sun2021loftr, edstedt2024roma}, we use a RANSAC threshold of 0.5 for pose estimation. The performance is reported as AUC at angular thresholds of \(5^{\circ}\), \(10^{\circ}\), and \(20^{\circ}\). As shown in Table \ref{tab:megadepth}, our method (L2M) outperforms existing methods, demonstrating the strong performance of our model in handling fine-grained details and complex geometric relationships.

\noindent\textbf{Cross-modal Generalization.} As shown in Table \ref{tab:Real_IR_pose}, we evaluate the zero-shot performance of our model, L2M, on the RGB-IR dataset (METU-VisTIR \cite{tuzcuouglu2024xoftr}), where all methods are trained solely on RGB data. Our method outperforms existing techniques across all error thresholds. Specifically, L2M achieves an AUC of \(30.13\%\) at \(5^\circ\), \(53.11\%\) at \(10^\circ\), and \(71.80\%\) at \(20^\circ\), demonstrating superior pose estimation accuracy compared to both sparse and dense matching methods. Besides, the dense matching methods, including DKM and GIM, demonstrate higher pose estimation accuracy, with DKM achieving \(22.53\%\) at \(20^\circ\). However, even the best-performing dense method, RoMa, with \(68.37\%\) at \(20^\circ\), remains substantially below the performance of L2M. These results highlight the robustness and effectiveness of our method, particularly in the challenging RGB-IR domain, where cross-modal matching is more complex and prone to large pose estimation errors.

\begin{figure}[t]
    \centering
    \subfloat[Paired Image]{\includegraphics[height=1.8in]{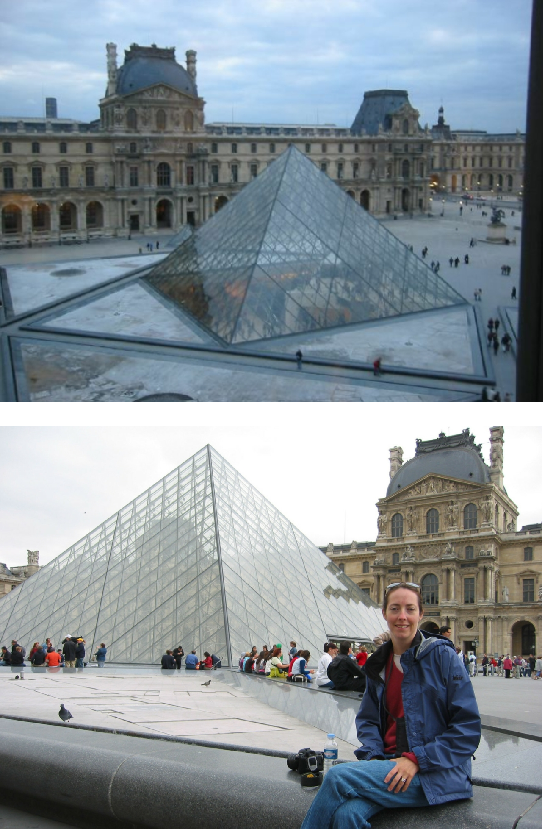}}
    \hfil
    \subfloat[w/o]{\includegraphics[height=1.8in]{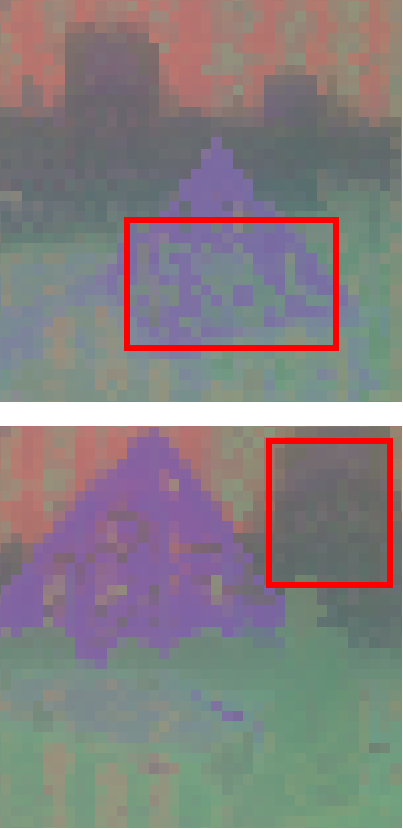}}
    \hfil
    \subfloat[\textbf{w/}]{\includegraphics[height=1.8in]{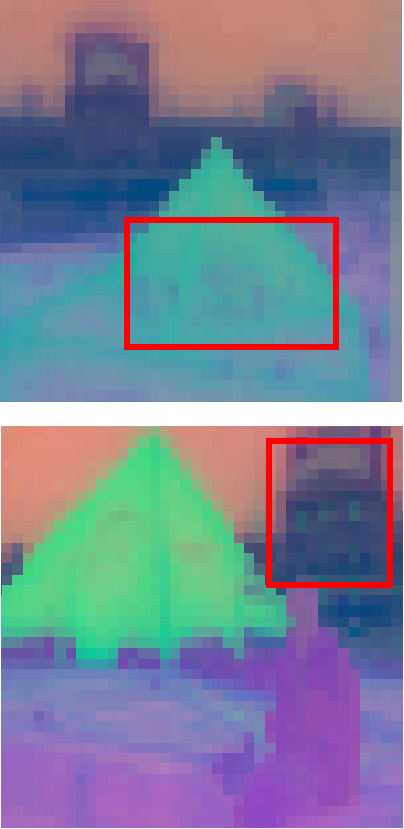}}
    \caption{Comparison of feature representations with and without the proposed 3D-aware encoder learning process. In these cases, the encoder after 3D-aware learning can establish detailed and meaningful correspondences.}
    \label{fig:vis-feature}
\end{figure}

\noindent\textbf{Qualitative Results.} As shown in Figure \ref{fig:visualization_warp}, we present qualitative results to demonstrate the effectiveness of our method compared to existing dense matching methods, particularly in challenging real-world and synthetic scenarios. Our method achieves denser matches in real-world scenes, almost achieving point-to-point correspondence. This is in stark contrast to state-of-the-art dense matching methods, which struggle to establish such precise correspondences. Our method is able to find detailed matches in complex environments, making it robust for practical applications.

\subsection{Discussions}

\noindent\textbf{Effectiveness of the 3D-aware encoder.} As shown in Table \ref{tab:ablations}, we conduct an ablation study to assess the contributions of key components. Specifically, we first evaluate the impact of incorporating a 3D-aware encoder (Stage 1). The results indicate that adding the 3D-aware encoder provides consistent improvements across both real-world and synthetic datasets.This highlights the importance of incorporating 3D-awareness in achieving robust feature matching performance across different domains.

\noindent\textbf{Effectiveness of the Robust Decoder Learning Process.} To further investigate the importance of our training strategy, we assess the effect of utilizing large-scale and diverse synthetic data for training the feature matching decoder (Stage 2). For comparison, we instead train the decoder using the MegaDepth dataset \cite{li2018megadepth}. The results show that the use of synthetic data is beneficial for improving generalization on datasets with limited real-world training data. This demonstrates the value of our data generation pipeline in augmenting feature matching models and enhancing their generalization ability in various real-world scenarios.


\noindent\textbf{Feature Visualization.} Furthermore, as shown in Figure \ref{fig:vis-feature}, we demonstrate the features of our method when using the 3D-aware encoder. In these cases, the encoder without 3D-aware learning process fails to establish detailed and meaningful correspondences, resulting in mismatched keypoints. In contrast, our method successfully identifies accurate and fine-grained correspondences, even in the presence of significant visual differences, such as the lack of texture in the towers and discontinuous features on translucent surfaces.

\section{Conclusion}

In this paper, we introduced \textbf{L2M}, a novel two-stage framework that enhances dense feature matching by lifting single-view 2D images into 3D space. Our approach addresses the limitations of conventional 2D image-based methods, which are constrained by their reliance on limited multi-view datasets captured in controlled environments. In particular, L2M incorporates a 3D-aware encoder learning strategy, which utilizes synthesized multi-view images and guided by explicit 3D feature Gaussians. This process injects multi-view geometric awareness into the encoder, enhancing its ability to handle challenging scenarios. Besides, a robust feature decoder is trained using large-scale synthetic novel views along with a re-rendering strategy, further improving the robustness and generalization of the feature decoder across diverse domains. Extensive experiments across Various zero-shot benchmarks demonstrate that our proposed L2M achieves state-of-the-art generalization performance, outperforming existing methods in handling real-world conditions and unseen domains.

\section*{Acknowledgements}

This work was supported by the National Key R\&D Program of China (2022YFC3300704), the National Natural Science Foundation of China (62331006, 62171038, and 62088101), and the Fundamental Research Funds for the Central Universities.

{
    \small
    \bibliographystyle{ieeenat_fullname}
    \bibliography{main}
}


\end{document}